%% file: main.tex
\documentclass{article}
 
 \pdfoutput=1
\usepackage{times}
\usepackage{epsfig}
\usepackage{amsmath,amsbsy,amsfonts,amssymb,amsthm,units,bm}
\usepackage[mathscr]{eucal}
\usepackage{graphicx}
\usepackage{subcaption}
\usepackage{algorithm, algorithmic}
\usepackage{color,cases}
\usepackage{array}
\usepackage{xhfill}
\usepackage{nicefrac}       
\usepackage{microtype}      
\usepackage{booktabs}       
\usepackage{makecell}
\usepackage{subfiles}
\usepackage{textcomp}
\usepackage{booktabs}
\usepackage{xcolor}
\usepackage{multirow}

\usepackage{hyperref}

\include{notations}
\usepackage[accepted]{icml2018}

\icmltitlerunning{Sturm: Sparse Tubal-Regularized Multilinear Regression for fMRI}

\begin{document}

\twocolumn[
\icmltitle{Sturm: Sparse Tubal-Regularized Multilinear Regression for fMRI}



\icmlsetsymbol{equal}{*}

\begin{icmlauthorlist}
\icmlauthor{Wenwen Li}{equal,Sheffield}
\icmlauthor{Jian Lou}{equal,HK}
\icmlauthor{Shuo Zhou}{Sheffield}
\icmlauthor{Haiping Lu}{Sheffield}
\end{icmlauthorlist}

\icmlaffiliation{Sheffield}{Department of Computer Science, University of Sheffield, UK}
\icmlaffiliation{HK}{Department of Computer Science, Hong Kong Baptist University, China}

\icmlcorrespondingauthor{Wenwen Li}{wenwen.li@sheffield.ac.uk}
\icmlcorrespondingauthor{Jian Lou}{loujian0425@gmail.com}
\icmlcorrespondingauthor{Shuo Zhou}{szhou20@sheffield.ac.uk}
\icmlcorrespondingauthor{Haiping Lu}{h.lu@sheffield.ac.uk}


\vskip 0.3in
]



\printAffiliationsAndNotice{\icmlEqualContribution} 

\begin{abstract}
While functional magnetic resonance imaging (fMRI) is important for healthcare/neuroscience applications, it is challenging to classify or interpret due to its multi-dimensional structure, high dimensionality, and small number of samples available. Recent sparse multilinear regression methods based on tensor are emerging as promising solutions for fMRI, yet existing works rely on unfolding/folding operations and a tensor rank relaxation with limited tightness. The newly proposed tensor singular value decomposition (t-SVD) sheds light on new directions. In this work, we study t-SVD for sparse multilinear regression and propose a \textbf{S}parse \textbf{tu}bal-\textbf{r}egularized \textbf{m}ultilinear regression (\textbf{Sturm}) method for fMRI. Specifically, the Sturm model performs multilinear regression with two regularization terms: a tubal tensor nuclear norm based on t-SVD and a standard $\ell_1$ norm. We further derive the algorithm under the alternating direction method of multipliers framework. We perform experiments on four classification problems, including both resting-state fMRI for disease diagnosis and task-based fMRI for neural decoding. The results show the superior performance of Sturm in classifying fMRI using just a small number of voxels.
\end{abstract}

\section{Introduction}
Brain diseases affect millions of people worldwide and impose significant challenges to healthcare systems. Functional magnetic resonance imaging (fMRI) is a key medical imaging technique for diagnosis, monitoring and treatment of brain diseases. Beyond healthcare, fMRI is also an indispensable tool in neuroscience studies \cite{faro2010bold}.

\begin{figure}[t]
\centering \makebox[0in]{
    \begin{tabular}{c c}
 \end{tabular}}
 \includegraphics[clip, trim=0.0cm 15cm 0.0cm 3.5cm, width=0.95\linewidth]{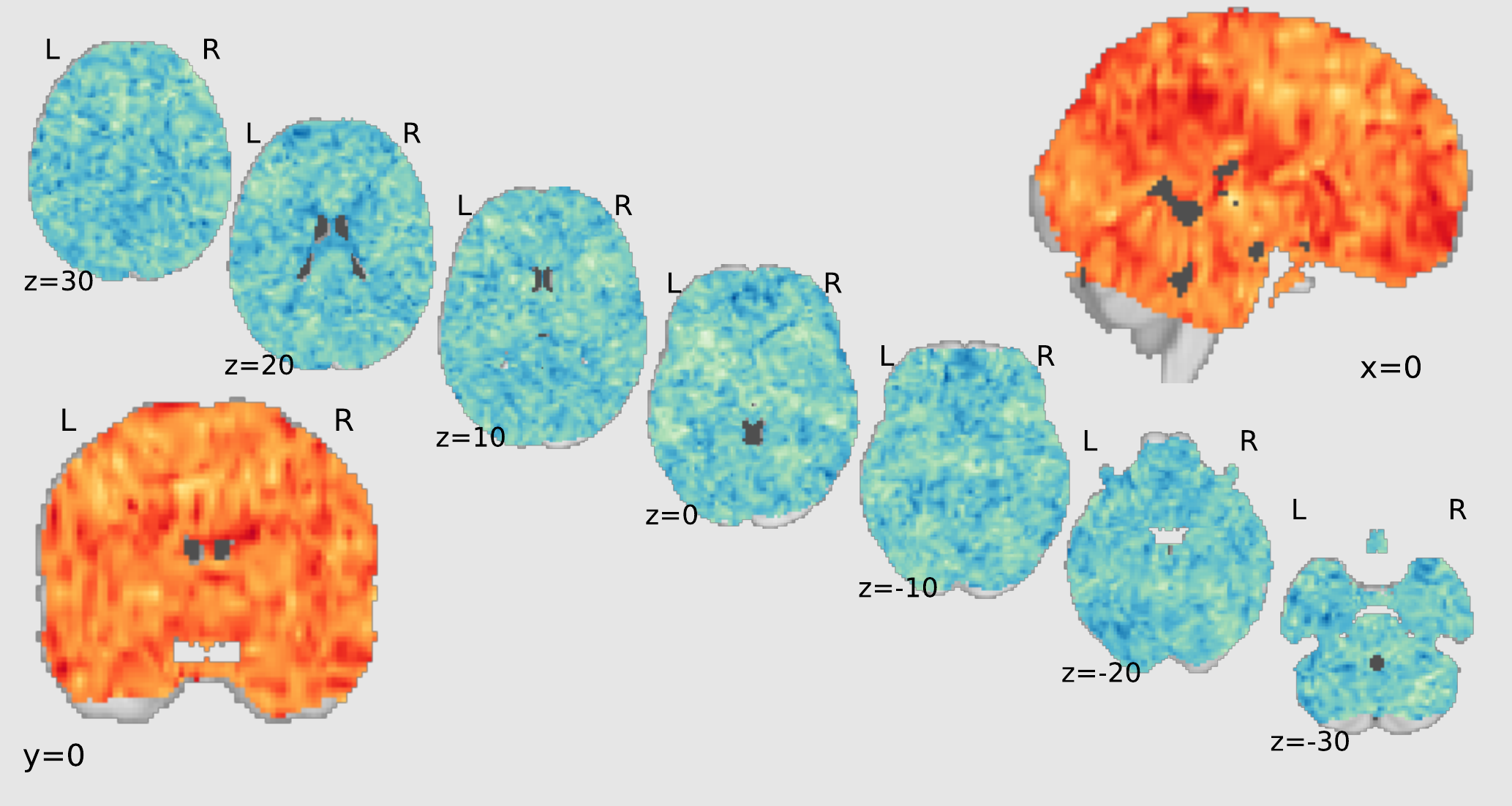}
  \caption{A 3D fMRI volume has three directions: sagittal (x), coronal (y) and axial (z). The original scans are taken with slices oriented in parallel to axial plane (as shown in the diagonal). We aim to classify fMRI with only a small number of voxels for easy interpretation by proposing a sparse multilinear regression model with tubal tensor nuclear norm regularization. (This figure is best viewed on screen.) }
  \label{fig:fMRI}
\end{figure}

fMRI records the Blood Oxygenation Level Dependent (BOLD) signal caused by changes in blood flow \cite{ogawa1990brain}, as depicted in Fig. \ref{fig:fMRI}. fMRI scan is a four-dimensional (4-D) sequence composed of magnetic resonance imaging (MRI) volumes sampled every few seconds, with over 100,000 voxels each MRI volume. It is often represented as a three-dimensonal (3-D) volume, with the statistics of each voxel summarized along the time axis. Figure \ref{fig:fMRI} visualizes example values of such statistics, showing very different characteristics from natural images. 

Unlike natural images, fMRI data is expensive to obtain. Thus, the number of fMRI samples in a study is typically limited to dozens. This makes fMRI challenging to analyze, particularly in its full (i.e., whole brain). Moreover, in healthcare and neuroscience, prediction accuracy is not the only concern. It is also important to interpret the learned features to domain experts such as clinicians or neuroscientists. This makes sparse learning models \cite{ryali2010sparse,simon2013sparse,rao2013sparse,hastie2015statistical} attractive because they can reveal direct dependency of a response with a small portion of input features. Therefore, tensor-based, sparse multilinear regression methods are emerging recently, where \textit{tensor} refers to multidimensional array. For simplicity, we consider only \textit{third-order} (i.e., 3-D) tensor in this paper.

Sparse multilinear regression models relate a predictor/feature tensor with a univariate response via a coefficient tensor, generalizing Lasso-based models  \cite{tibshirani1996regression,hastie2015statistical} to tensor data. Regularization that promotes sparsity and low rankness is also generalized to the coefficient tensor. For example, the regularized multilinear regression and selection (Remurs) model \cite{song2017multilinear} incorporates a sparse regularization term, via an $\ell_1$ norm,  and a \textit{Tucker rank}-minimization term, via a summation of the nuclear norms (SNN) of unfolded matrices. There are also CANDECOMP/PARAFAC (CP) rank-based methods \cite{tan2012logistic,he2018boosted}. For example, the fast stagewise unit-rank tensor factorization (SURF) \cite{he2018boosted} enforces the low rankness via the CP decomposition model \cite{hitchcock1927expression,harshman1970foundations,carroll1970analysis}, with the rank to be incrementally estimated via residual computation rather than direct minimization. And it proposes a divide-and-conquer strategy to improve efficiency and scalability with a greedy algorithm \cite{cormen2009introduction}. Although Remurs minimizes the Tucker rank directly, it involves unfolding/folding operations that transform tensor to/from matrix, which break some tensor structure, and SNN is not a tight convex relaxation of its target rank \cite{romera2013new}.

A new Tubal Tensor Nuclear Norm (TNN) \cite{zhang2014novel,zhang2017exact,lu2016tensor,lu2018tensor} is recently proposed based on the \textit{tubal rank}, which originates from the tensor singular value decomposition (t-SVD) \cite{kilmer2013third}. 
In tensor recovery problems, TNN guarantees the exact recovery under certain conditions. Improved performance has been observed on image/video data on tensor completion/recovery and robust tensor principal component analysis (PCA)  \cite{zhang2017exact,lu2016tensor,lu2018tensor}. In addition, TNN does not require unfolding/folding of tensor in its optimization.

In this work, we study sparse multilinear regression under the t-SVD framework for fMRI classification. The success of TNN is limited to \textit{unsupervised} learning settings such as completion/recovery and robust PCA. To our knowledge, TNN has not been studied in a \textit{supervised} setting yet, such as multilinear regression where the problem is not recovery given samples of a tensor but prediction of a response with a set of training tensor samples. Moreover, the targeted fMRI classification tasks have additional challenge of small sample size (relative to the feature dimension).

Our contributions are twofold: Firstly, we propose a \textbf{S}parse \textbf{tu}bal-\textbf{r}egularized \textbf{m}ultilinear regression (\textbf{Sturm}) method that incorporates TNN regularization. Specifically, we formulate the \textit{Sturm} model by incorporating both a TNN regularization and a sparsity regularization on the coefficient tensor. We solve the resulted \textit{Sturm} problem using the alternating direction method of multipliers (ADMM) framework \cite{boyd2011distributed}. TNN-based formulation allows efficient parameter update in the Fourier domain, which is highly parallelizable. 

Our second contribution is that we evaluate Sturm and related methods on \textit{both} resting-state and task-based fMRI classification problems, instead of only one of them as in previous works \cite{zhou2013discriminative, zhou2014discriminative, shi2014joint, zhou2017revisiting, he2017multi}. We use public datasets with identifiable subsets for repeatibility, and examine both the classification accuracy and sparsity. The results show Sturm outperforming other state-of-the-art methods on the whole.

\section{Related Work}
\textbf{Tensor decomposition and rank.} There are three popular tensor decomposition methods and associated tensor rank definitions: \textbf{1)} The CP decomposition of a tensor is written as the summation of $R$ \textit{rank-one} tensors \cite{hitchcock1927expression,harshman1970foundations,carroll1970analysis}, where $R$ is the CP rank. \textbf{2)} For a 3-D tensor, the Tucker decomposition decomposes it into a core tensor and three factor matrices \cite{de2000multilinear}, and the \textit{Tucker rank} is 3-tuple consisting of the rank for each mode-$n$ unfolded matrix. \textbf{3)} The t-SVD views a 3-D tensor as a matrix of \textit{tubes} (mode-$3$ vectors) oriented along the third dimension and decomposes it as circular convolutions on three tensors \cite{braman2010third, kilmer2013third,kilmer2011factorization, gleich2013power}.  This leads to a new tensor \textit{tubal rank} defined as the number of non-zero singular tubes. Please refer to \cite{zhang2017exact,zhang2014novel,yuan2016tensor,lu2016tensor,lu2018tensor,mu2014square} for more detailed discussion, e.g., on their pros and cons.

\textbf{Low-rank tensor completion/recovery.} Tensor completion/recovery takes a tensor with missing/noisy entries as the input and aims to recover/complete those entries. Low-rank assumption makes solving such problems feasible, and often with provable theoretical guarantees under mild conditions. The three types of tensor decomposition have their corresponding tensor completion/recovery approaches that minimize respective tensor ranks. Direct minimization of tensor rank is NP-hard \cite{hillar2013most}. Therefore, CP rank, Tucker rank, and tubal rank are relaxed to CP-based nuclear norm \cite{shi2017tensor},  the sum of matrix nuclear norm   \cite{liu2013tensor}, and the tubal tensor nuclear norm.

\textbf{Sparse and low-rank multilinear regression.} Sparse and low-rank constraints have also been applied to multilinear regression problems. Multilinear regression models \cite{signoretto2010nuclear,su2012multivariate,guo2012tensor,zhou2013tensor} relate a predictor/feature tensor $\tX \in\mathbb{R}^{I_1\times I_2\times I_3}$ with a univariate response $y$, via a coefficient tensor $\tW \in \mathbb{R}^{I_1\times I_2\times I_3}$. The Remurs model \cite{song2017multilinear} introduces regularization with a sparsity constraint, via $\ell_1$ norm, and a low-rank constraint via Tucker rank-based SNN. Instead of minimizing the rank, the fast stagewise unit-rank tensor factorization (SURF) \cite{he2018boosted} is an efficient and scalable method that imposes a low CP rank constraint by setting a maximum rank and increasing the rank estimate from 1 to the maximum, stopping upon zero residual.

\textbf{Tensor methods on fMRI.} Due to the high dimensionality of fMRI data, regions of interest (ROIs) are typically used rather than all voxels in the original 3-D spatial domain \cite{chen2015reduced,he2018boosted}. ROI analysis requires strong prior knowledge to determine the regions. In contrast, whole-brain fMRI analysis \cite{poldrack2013toward} is more data-driven. Thus, tensor-based machine learning methods \cite{cichocki2013tensor,cichocki2009nonnegative} have been developed for fMRI, with promising results reported \cite{acar2017tensor, zhou2013tensor, song2015learning, he2017multi, song2017multilinear,ozdemir2017multi,barnathan2011twave}. However, in these works, the learning methods are only evaluated on either resting-state fMRI for disease diagnosis \cite{zhou2017revisiting} or task-based fMRI for neural decoding \cite{he2017multi,chen2015reduced,song2017multilinear}, but not both.

\section{Tubal Tensor Nuclear Norm}

\begin{table}[t]
\begin{small}
	\centering
	\begin{tabular}{ m{1cm} m{6cm} } 
		\toprule
		\textbf{Notation} & \textbf{Description} \\
		\midrule
		$a$ & Lowercase letter denotes scalar \\
		$ \mathbf{a} $ & Bold lowercase letter denotes vector \\
		$ \mathbf{A} $  & Bold uppercase letter denotes matrix \\
		$ \tA $ & Calligraphic uppercase letter denotes tensor \\
		$\tA * \tB $ & t-product between tensors $\tA$ and $\tB$\\
		$\tA^{\top}$ & Tensor conjugate transpose of $\tA$\\
		$\mA^{(i_3)}$ & The $i_3$th mode-3 (frontal) slice of $\tA$\\
		$\ftA$ & The discrete Fourier transform of $\tA$ \\
		\bottomrule	
	\end{tabular}
\end{small}	
	\caption{Important notations.}
	\label{tab:notations}	
\end{table}

\textbf{Notations.} Table \ref{tab:notations} summarizes important notations used in this paper. We use lowercase, bold lowercase, bold uppercase, calligraphic uppercase letters to denote scalar, vector, matrix, and tensor, respectively. We denote indices by lowercase letters spanning the range from 1 to the uppercase letter of the index, e.g., $m =1,\ldots,M$. A third-order tensor $\tA\in \mathbb{R}^{I_1\times I_2 \times I_3}$ is addressed by three indices $\{i_n\}$, $n=1, 2, 3$. Each $i_n$ usually addresses the $n$th mode of $\tA$, while such convention may not be strictly followed when the context is clear. The $i_3$th mode-3 slice, a.k.a. the frontal slice, of $\tA$ is denoted as $\mA^{(i_3)}$, a matrix obtained by fixing the mode-3 index $i_3$, i.e., $\mA^{(i_3)}=\tA(:,:,i_3)$. The $(i_1,i_2)$th \textit{tube} of $\tA$, denoted as $\mathring{\vect{a}}_{i_1i_2}$, is a mode-$3$ vector obtained by fixing the first two mode indices, i.e., $\tA(i_1,i_2,:)$.

\begin{figure}[t]
\centering \makebox[0in]{
    \begin{tabular}{c c}
      \includegraphics[scale=0.24]{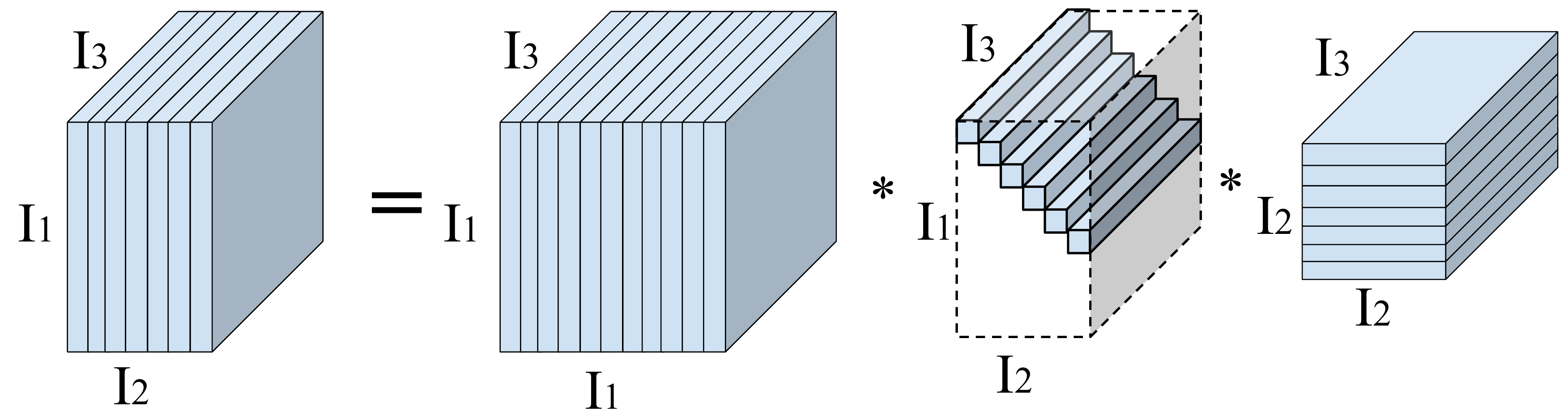}
 \end{tabular}}
  \caption{Illustration of t-SVD $\tA = \tU*\tS*\tV^{\top}$, assuming $I_1>I_2$.}
  \label{fig:tSVD}
\end{figure}

\textbf{t-SVD and tubal rank.} We first review the t-SVD framework following the definitions in \cite{kilmer2013third}. The \textbf{t-product} (\textit{tensor-tensor product}) between tensor $\tA\in\mathbb{R}^{I_1\times I_2\times I_3}$ and $\tB\in\mathbb{R}^{I_2\times J_4\times I_3}$ is defined as $ \tA * \tB = \tens{C}\in\mathbb{R}^{I_1\times J_4\times I_3}$. The $(i_1,j_4)$th tube $\mathring{\vect{c}}_{i_1j_4}$ of $\tens{C}$ is computed as
  \begin{equation}
  \label{Eq.fiber_prod}
    \mathring{\vect{c}}_{i_1j_4} = \tens{C}(i_1,j_4,:) = \sum _{i_2=1}^{I_2}\tA(i_1,i_2,:)*\tB(i_2,j_4,:),
  \end{equation}
where $*$ denotes the circular convolution \cite{rabiner1975theory} between two tubes (vectors) of the same size and the respective t-product between tensors. The \textbf{tensor conjugate transpose} of $\tA\in\mathbb{R}^{I_1\times I_2\times I_3}$ is denoted as $\tA^{\top}\in\mathbb{R}^{I_2 \times I_1\times I_3}$, obtained by conjugate transposing each of the frontal slice and then reversing the order of transposed frontal slices $\tA(:,:,i_3)$. A tensor $\tens{I}\in\mathbb{R}^{I\times I\times I_3}$ is an \textbf{identity tensor} if its first mode-3 slice  $\tens{I}^{(1)}$ is an $I\times I$ identity matrix and all the rest mode-3 slices, i.e. $\tens{I}^{(i_3)}$ for $i_3=2,...,I_3$, are zero matrices. An \textbf{orthogonal tensor} is a tensor $\tens{Q}\in\mathbb{R}^{I\times I\times I_3}$ that satisfies the following condition,
  \begin{equation}
    \tens{Q}^{\top}*\tens{Q} = \tens{Q}*\tens{Q}^{\top} = \tens{I},
  \end{equation}
  where $\tens{I}\in\mathbb{R}^{I\times I\times I_3}$ is an identity tensor and $*$ is the t-product defined above. If $\tA$'s all mode-3 slices $\mA^{(i_3)}$, $i_3=1,...,I_3$ are diagonal matrices, it is called an \textbf{f-diagonal tensor}. Based on these definitions, the \textbf{t-SVD} of $\tA \in \R^{I_1\times I_2\times I_3}$ is defined as
  \begin{equation}
  \label{Eq.tsvd}
    \tA = \tU * \tS * \tV^{\top},
  \end{equation}
 where $\tU\in\mathbb{R}^{I_1\times I_1\times I_3}$, $\tV\in\mathbb{R}^{I_2\times I_2\times I_3}$ are orthogonal tensors, and $\tS\in\mathbb{R}^{I_1\times I_2\times I_3}$ is an f-diagonal tensor.
This t-SVD definition leads to a new tensor rank, the \textbf{tubal rank}, which is defined as the number of nonzero singular tubes of $\tS$, i.e. $\#\{i:\tS(i_2,i_2,:)\neq 0\}$, assuming $I_1\geq I_2$. Figure \ref{fig:tSVD} is an illustration of t-SVD.

\textbf{t-SVD via Fourier transform.} t-SVD can be computed via the discrete Fourier transform (DFT) for better efficiency. We denote the Fourier transformed tensor $\tA$ as $\ftA$, obtained via fast Fourier transform (FFT) along mode-3, i.e., $\ftA = {\tt fft}(\tA,[\hspace{1mm}],3)$. The connection between t-SVD and DFT is detailed in \cite{kilmer2013third}.

\textbf{Tubal Tensor nuclear norm.} TNN is a convex relaxation for the tubal rank, as an average tubal multi-rank within the unit tensor spectral norm ball \cite{lu2018tensor}. It can be defined via DFT, similar to t-SVD computation above. The TNN of $\tA \in \R^{I_1\times I_2\times I_3}$ is defined as
\begin{equation}
   \|\tA\|_{TNN} = \frac{1}{I_3}\sum _{i_3=1}^{I_3} \|\fmA^{(i_3)}\|_*,
\end{equation}
where $\|\cdot \|_* $ denotes the matrix nuclear norm. Please refer to \cite{lu2018tensor} for a detailed derivation and complete theoretical analysis, e.g., on the tightness of the relaxation.

Note that when $I_3=1$, the above definitions for tensors will be equivalent to the counterparts for matrices.

\section{Sparse Multilinear Regression with Tubal Rank Regularization}

Tubal rank-based TNN has shown to be superior to Tucker rank-based SNN in tensor completion/recovery \cite{zhang2017exact}, and tensor robust PCA \cite{lu2016tensor}, which are all unsupervised learning settings. To our knowledge, there is no study on TNN in a supervised learning setting yet. In this work, we explore the supervised learning with TNN and study whether TNN can improve supervised learning, e.g., multilinear regression. In the following, we propose the Sturm model and derive the Sturm algorithm under the ADMM framework.

\subsection{The Sturm model}
We incorporate TNN in the multilinear regression problem, which trains a model from $M$ pairs of feature tensors and their response labels $(\tX_m \in \mathbb{R}^{I_1\times I_2\times I_3}, y_m)$ with $m = 1,...,M$ to relate them via a \textit{coefficient tensor} $\tW \in \mathbb{R}^{I_1\times I_2\times I_3}$. This can be achieved by minimizing a loss function, typically with certain regularization:
\begin{equation} 
\label{eq.mlr.general_def}
	\min _{\tW} \frac{1}{M} \sum _{m=1}^M L(\la \tX_m,\tW\ra, y_m) + \lambda \Omega(\tW),
\end{equation}
where $L(\cdot)$ is a loss function, $\Omega(\cdot)$ is a regularization function, $\lambda$ is a balancing hyperparameter, and $\la \tX,\tW\ra$ denotes the inner product (a.k.a. the scalar product) of two tensors of the same size defined as
\begin{align}\label{innerprodTensor}
  \langle \tX, \tW\rangle
  :=\sum_{i_1}\sum_{i_2}\sum_{i_3}
  \tX(i_1,i_2,i_3)\cdot \tW(i_1,i_2,i_3),
\end{align} 
 
The Remurs \cite{song2017multilinear} model uses a conventional least square loss function and assumes $\tW$ to be both sparse and low rank. The sparsity of $\tW$ is regularized by an $\ell _1$ norm and the rank by an SNN norm. However, the SNN requires unfolding $\tW$ into matrices, susceptible to losing some higher-order structural information. Moreover, it has been pointed out in \cite{romera2013new} that SNN is not a tight convex relaxation of its target rank.

This motivates us to propose a \textbf{S}parse \textbf{tu}bal-\textbf{r}egularized \textbf{m}ultilinear regression (\textbf{Sturm}) model which replaces SNN in Remurs with TNN. This leads to the following objective function
\begin{equation}
\label{eq.sturm.formulation}
   \min _{\tW} \frac{1}{2} \sum _{m=1}^M (y_m-\la \tX_m,\tW \ra)^2 + \tau \|\tW\|_{TNN} + \gamma\|\tW\|_{1},
\end{equation}
where $\tau$ and $\gamma$ are hyperparameters, and $\|\tW\|_{1}$ is the $\ell_1$ norm of tensor $\tW$, defined as 
\begin{equation}\label{eqn:l1}
\|\tW\|_1
  = \sum_{i_1}\sum_{i_2}\sum _{i_3}
  \left|\tW(i_1,i_2,i_3)\right|,
\end{equation}
which is equivalent to the $\ell_1$ norm of its vectorized representation $\vect{w}$. Here, the TNN regularization term $\|\tW\|_{TNN}$ enforces low tubal rank in $\tW$. The trade-off between $\tau$ and $\gamma$ as well as the degenerated versions follow the analysis for the Remurs \cite{song2017multilinear}.

\subsection{The Sturm algorithm via ADMM}
ADMM \cite{boyd2011distributed} is a standard solver for Problem (\ref{eq.sturm.formulation}). Thus, we derive an ADMM algorithm to optimize the Sturm objective function. We begin with introducing two auxiliary variables, $\tA$ and $\tB$ to disentangle the TNN and the $\ell _1$-norm regularization:
\begin{gather}
    \min _{\tW} \frac{1}{2} \sum _{m=1}^M (y_m-\la \tX_m,\tA \ra)^2 + \tau \|\tB\|_{TNN} + \gamma\|\tW\|_{1} \\
    s.t.\ \tA = \tW\ \text{and } \tB = \tW. \nonumber
\end{gather}
Then, we introduce two Lagrangian dual variables $\tP$ (for $\tA$) and $\tQ$ (for $\tB$). With a Lagrangian constant $\rho$, the augmented Lagrangian becomes,
 \begin{equation}
     \begin{split}
    & L_{\rho}(\tA,\tB,\tW,\tP,\tQ)  =  \frac{1}{2} \sum _{m=1}^M (y_m-\la \tX_m,\tA \ra)^2\\
     & + \tau \|\tB\|_{TNN} + \gamma\|\tW\|_{1}\\
     & + \Big \la \tP, \tA-\tW \Big \ra + \frac{\rho}{2}\|\tA-\tW\|_F^2 \\ & +  \Big \la \tQ, \tB-\tW \Big \ra + \frac{\rho}{2}\|\tB-\tW\|_F^2.
     \end{split}   
  \end{equation}
We further introduce two scaled dual variables $\tens{P'} = \frac{1}{\rho}\tP$ and $\tens{Q'} = \frac{1}{\rho}\tQ$ only for notation convenience. Next, we derive the update from iteration $k$ to $k+1$ by taking an alternating strategy, i.e., minimizing one variable with all other variables fixed.

\noindent\textbf{Updating $\tA ^{k+1}$:} 
\begin{equation}\label{eq:l1normUpdateA}
  \begin{split}
  &\tA ^{k+1} = \arg\min _{\tA} L_{\rho} (\tA,\tB ^{k},\tW ^{k},\tens{P'} ^{k},\tens{Q'} ^{k})\\
              & =  \arg\min _{\tA} \frac{1}{2} \sum _{m=1}^M (y_m-\la \tX_m,\tA \ra)^2 + \frac{\rho}{2} \|\tA-\tW^k + \tens{P'}^{k}\|_F.
  \end{split}
 \end{equation}

This can be rewritten as a linear-quadratic objective function by vectorizing all the tensors. Specifically, let $\vect{a}={\tt vec}(\tA)$, $\vect{w}^k={\tt vec}(\tW ^k)$,  $\vect{p'}^k={\tt vec}(\tens{P'} ^k)$, $\vect{y}=[y_1 \cdots y_M]^\top$, $\vect{x}_m={\tt vec}(\tX_m)$, and  $\mat{X}=[\mathbf{x}_1 \cdots \mathbf{x}_M]^\top$. Then we get an equivalent objective function with the following solution: 
\begin{equation}\label{eq:updateA}
\vect{a}^{k+1} = (\X^\top\X+\rho\I)^{-1}(\X^\top\y + \rho(\w^k - \mathbf{p'}^{k})),
\end{equation}
where $\I$ is an identity matrix. Note that this does not break/lose any structure because Eq. (\ref{eq:l1normUpdateA}) and Eq. (\ref{eq:updateA}) are equivalent. $\tA^{k+1}$ is obtained by folding (reshaping) $\vect{a}^{k+1}$ into a third-order tensor, denoted as $\tA^{k+1} = {\tt tensor}_{3}(\vect{a}^{k+1})$. 
Here, for a fixed $\rho$, we can avoid high per-iteration complexity of updating $\vect{a}^{k+1}$ by pre-computing a Cholesky decomposition of $(\X^\top\X+\rho\I)$, which does not change over iterations. 

\noindent\textbf{Updating $\tB ^{k+1}$:}
\begin{equation}
   \begin{split}
    &\tB ^{k+1} = \arg\min _{\tB} L_{\rho} (\tA^{k+1},\tB,\tW ^{k},\tens{P'} ^{k},\tens{Q'} ^{k})\\
    & = \arg\min _{\tB} \tau \|\tB\|_{TNN} + \frac{\rho}{2} \|\tB-\tW^k+\tens{Q'}^{k}\|_F^2 \\
    & ={\tt prox}_{\frac{\tau}{\rho}\|\cdot\|_{TNN}} (\tW^k-\tens{Q'}^{k}).
   \end{split}  
\end{equation}

This means that $\tB ^{k+1}$ can be solved by passing parameter $\frac{\tau}{\rho}$ to the proximal operator of the TNN \cite{zhang2014novel,zhang2017exact}. The proximal operator for the TNN at tensor $\tT$ with parameter $\mu$ is denoted by ${\tt prox}_{\mu \|\cdot\|_{TNN}} (\tT)$ and defined as 
\begin{equation}
   {\tt prox}_{\mu \|\cdot\|_{TNN}} (\tT) := \arg\min _{\tW} \mu\|\tW\|_{TNN} + \frac{1}{2} \|\tW-\tT\|_F^2,
\end{equation}
where $\|\cdot\|_F$ is the Frobenius norm defined as $\|\tT\|_F = \sqrt{\la \tT, \tT \ra}$ using Eq. (\ref{innerprodTensor}).
The proximal operator for TNN can be more efficiently computed in the Fourier domain, as in Algorithm \ref{alg:prox_TNN}, where in Step \ref{lst:line:threshold}, $(\vect{s}-\mu)_{+}=\max\{\vect{s}-\mu,0\}$ and ${\tt diag}(\mathbf{a})$ denotes a diagonal matrix whose diagonal elements are from $\mathbf{a}$.

\begin{algorithm}[t]
  \caption{Proximal Operator for TNN: ${\tt prox}_{\mu \|\cdot\|_{TNN}} (\tT)$}
  \begin{algorithmic}[1]
  \REQUIRE $\tT \in \mathbb{R}^{I_1 \times I_2\times I_3}, \mu$
    \STATE $\ftT = {\tt fft}(\tT,[\hspace{1mm}],3)$; 
  \FOR{$i_3 = 1,2,...,I_3$}
    \STATE $ [\U, {\tt diag}(\vect{s}), \V] = {\tt svd}(\fmT^{(i_3)})$
    \STATE $ \fmZ^{(i_3)} = \U({\tt diag}((\vect{s}-\mu)_{+}))\V^{\top};$\label{lst:line:threshold}
  \ENDFOR
        \STATE ${\tt prox}_{\mu \|\cdot\|_{TNN}} (\tT) = {\tt ifft}(\ftZ,[\hspace{1mm}],3)$;
 \ENSURE ${\tt prox}_{\mu \|\cdot\|_{TNN}} (\tT)$ 
  \end{algorithmic}
  \label{alg:prox_TNN}
\end{algorithm}

\noindent\textbf{Updating $\tW ^{k+1}$:}
\begin{equation}
   \begin{split}
    &\tW ^{k+1} = \arg\min _{\tW} L_{\rho} (\tA^{k+1},\tB ^{k+1}, \tW, \tens{P'}^{k},\tens{Q'}^{k})\\
    & ={\tt prox}_{\frac{\gamma}{2\rho}\|\cdot\|_{1}}  \Big(\frac{\tA^{k+1}+\tens{P'}^{k} + \tB^{k+1}+\tens{Q'}^{k}}{2} \Big).
   \end{split}
\end{equation} 

It can be solved by calling the proximal operator of the $\ell _1$ norm with parameter $\frac{\gamma}{2\rho}$, which is simply the element-wise soft-thresholding, i.e. 
\begin{equation}\label{eq:l1prox}
{\tt prox}_{\mu \|\cdot\|_1}(\tT) = (\tT-\mu)_{+}.
\end{equation}

\noindent\textbf{Updating $\tP ^{k+1}$ and $\tQ^{k+1}$:} 
The updates of $\tP$ and $\tQ$ are simply dual ascent steps: 
\begin{gather}
   \tens{P'}^{k+1} = \tens{P'}^{k} + \tA^{k+1}-\tW^{k+1},\\
   \tens{Q'}^{k+1} = \tens{Q'}^{k} + \tB^{k+1}-\tW^{k+1}.
\end{gather}
The complete procedure is summarized in Algorithm \ref{alg:ADMM_Sturm}. The code will be made publicly available via GitHub.

\begin{algorithm}[t]
  \caption{ADMM for \textbf{Sturm}}
  \begin{algorithmic}[1]
  \REQUIRE $(\tX _m, y_m)$ for $m = 1,...,M$, $\tau$, and $\lambda$;
    \STATE Initialize $\tA^0, \tB^0, \tW^0, \tP'^0, \tQ'^0$ to all zero-tensors and set $\rho$ and $K$.
  \FOR{$k = 1,...,K$}
    \STATE Update $\tA^{k+1}$ by Eq. (\ref{eq:updateA});\label{lst:line:updatetA}
    \STATE Update $\tB^{k+1}$ by Alg. \ref{alg:prox_TNN} as ${\tt prox}_{\frac{\tau}{\rho}\|\cdot\|_{TNN}} (\tW^k-\tens{Q'}^{k})$;\label{lst:line:updatetB}
    \vspace{-0.2cm}
    \STATE Update $\tW^{k+1}$ by Eq. (\ref{eq:l1prox}) as
    \vspace{-0.2cm}
    $${\tt prox}_{\frac{\gamma}{2\rho}\|\cdot\|_{1}} \Big(\frac{\tA^{k+1}+\tens{P'}^{k} + \tB^{k+1}+\tens{Q'}^{k}}{2} \Big);$$\label{lst:line:updatetW}
    \vspace{-0.2cm}
    \STATE $\tens{P'}^{k+1} = \tens{P'}^{k} + \tA^{k+1}-\tW^{k+1}$;\label{lst:line:updatetP}
   	\STATE $\tens{Q'}^{k+1} = \tens{Q'}^{k} + \tB^{k+1}-\tW^{k+1}$;\label{lst:line:updatetQ}
  \ENDFOR
 \ENSURE $\tW^{K}$ 
  \end{algorithmic}
  \label{alg:ADMM_Sturm}
\end{algorithm}
 
\subsection{Computational complexity}
Finally, we analyze the per-iteration computational complexity of Algorithm \ref{alg:ADMM_Sturm}. Let $I=I_1I_2I_3$. Step~\ref{lst:line:updatetA} takes $O(IM + \min\{M^2, I^2\})$. Step ~\ref{lst:line:updatetB} takes $O(\min\{I_1,I_2\}I)$ for the singular value thresholding in the Fourier domain, plus $O(I\log(I_3))$ for ${\tt fft}$ and ${\tt ifft}$. Step~\ref{lst:line:updatetW} takes $O(I)$ because the proximal operation for $\ell _1$ norm is element-wise. Step~\ref{lst:line:updatetP} and Step~\ref{lst:line:updatetQ} take $O(I)$. As a result, in a high dimensional (or small sample) setting where $I \gg M$, the per-iteration complexity is $O(I(\log(I_3)+M))$.

\section{Experiments}
We evaluate our \textit{Sturm} algorithm on four binary classification problems with six datasets from three public fMRI repositories. While known works typically focus on either resting-state or task-based fMRI only and rarely both, here we study \textit{both} types. We test the efficacy of the Sturm approach against five state-of-the-art algorithms and three additional variations, in terms of both classification accuracy and sparsity.

\subsection{Classification problems and datasets}
\textit{Resting-state fMRI for disease diagnosis.} Resting-state fMRI is scanned when the subject is not doing anything, i.e., at rest. It is commonly used for brain disease diagnosis, i.e., classifying clinical population. In this paper, we consider only the binary classification of patients (positive) and health control subjects (negative).

\textit{Task-based fMRI for neural decoding.} Task-based fMRI is scanned when the subject is performing certain tasks, such as viewing pictures or reading sentences \cite{wang2003using}. It is commonly used for studies decoding brain cognitive states, or neural decoding. The objective is to classify (decode) the tasks performed by subjects using the fMRI information. In this paper, we consider only binary classification of two different tasks.

\textbf{Chosen datasets.} We study four fMRI classification problems on six datasets from three public repositories, with the key information summarized in Table \ref{tab:dataset}. Two are disease diagnosis problems on resting-state fMRI, and the other two are neural decoding problems on task-based fMRI, as described below.
\begin{itemize}
    \item \textbf{Resting 1 -- ABIDE\textsubscript{NYU\&UM}}: The Autism Brain Imaging Data Exchange (ABIDE)\footnote{\url{http://fcon_1000.projects.nitrc.org/indi/abide}} \cite{craddock2013neuro} consists of patients with autism spectrum disorder (ASD) and healthy control subjects. We chose the largest two subsets contributed by New York University (NYU) and University of Michigan (UM). The fMRI data has been preprocessed by the pipeline of Configurable Pipeline for the Analysis of Connectomes (CPAC). Quality control was performed by selecting the functional images with quality `OK' reported in the phenotype data. 
    \item \textbf{Resting 2 -- ADHD-200\textsubscript{NYU}}: We chose the NYU subset from the Attention Deficit Hyperactivity Disorder (ADHD) 200 (ADHD-200) dataset\footnote{\url{http://neurobureau.projects.nitrc.org/ADHD200/Data.html}} \cite{bellec2017neuro}, with ADHD patents and healthy controls. The raw data is preprocessed by the pipeline of Neuroimaging Analysis Kit (NIAK). 
    \item \textbf{Task 1 -- Balloon vs Mixed gamble }: We chose two gamble-related datasets from the OpenfMRI repository\footnote{Data used in this paper are available at OpenfMRI: \url{https://legacy.openfmri.org}, now known as OpenNeuro: \url{https://openneuro.org}.}  
    \cite{poldrack2013toward} project to form a classification problem. They are 1) Balloon analog risk-taking task (BART) and 2) Mixed gambles task.
    \item \textbf{Task 2 -- Simon vs Flanker}: We chose another two recognition and response related tasks from OpenfMRI for binary classification. They are 1) Simon task and 2)  Flanker task.
\end{itemize}

\textit{Resting-state fMRI preprocessing.} The raw resting-state brain fMRI data is 4-D. We follow typical approaches to reduce the 4-D data to 3-D by either taking the average \cite{he2017multi} or the amplitude \cite{yu2007altered} of low frequency fluctuation of voxel values along the time dimension. We perform experiments on both and report the best results. 

\textit{Task-based fMRI preprocessing.} Following \cite{poldrack2013toward}, we re-implemented a preprocessing pipeline to process the OpenfMRI data to obtain the 3D statistical parametric maps (SPMs) for each brain condition with a standard template. We used the same criteria as in \cite{poldrack2013toward} to selected one contrast (one specific brain condition over experimental conditions) per task for classification. 

\textit{The tubal mode of fMRI.} Figure \ref{fig:fMRI} illustrates how fMRI scan is obtained along the axial direction, which is mode 3 in tensor representation. Each image along the diagonal is a mode-3 (frontal) slice. Therefore, it is a natural choice to consider mode 3 as the tubal mode to apply Sturm.

\begin{table}[t]
\begin{small}
\setlength{\tabcolsep}{2pt}
 	\centering
 	\begin{tabular}{ m{3.8cm} m{0.8cm} m{0.9cm} m{2cm} } 
 		\toprule
 		\textbf{Classification Problem} & \textbf{{\# Pos.}} & \textbf{{\# Neg.}} & \textbf{Input data size} $I_1 \times I_2 \times I_3$ \\
 		\midrule
 		Resting 1 -- ABIDE\textsubscript{NYU\&UM} & 101 & 131 & $61 \times 73 \times 61$ \\
 		Resting 2 -- ADHD-200\textsubscript{NYU} & 118 & 98 & $53 \times 64 \times 46$ \\
 		Task 1 -- Balloon vs Mixed & 32 & 32 & $91 \times 109 \times 91$\\
 		Task 2 -- Simon vs Flanker & 42 & 52 & $91 \times 109 \times 91$\\
 		\bottomrule	
 	\end{tabular}
\end{small} 	
\caption{Summary of the four classification problems and respective datasets. \# denotes the number of volumes (samples). Pos. and Neg. are short for the positive and negative classes, respectively. For diagnosis problems on ABIDE/ADHD-200, patients and health subjects are considered as positive and negative classes, respectively. The two neural decoding problems are formed by using six OpenfMRI datasets listed as Pos. vs Neg.}
\label{tab:dataset}
\end{table}

\subsection{Algorithms}

We evaluate Sturm and Sturm + SVM (support vector machine) against the following five algorithms and three additional algorithms via combination with SVM.
\begin{itemize}
    \item \textit{SVM}: We chose linear SVM for both speed and prediction accuracy. (We studied both the linear and Gaussain RBF kernel SVM and found the linear one performs better on the whole.) 
    \item \textit{Lasso}: It is a linear regression method with the $\ell_1$ norm regularization.
    \item \textit{Elastic Net (ENet)}: It is a linear regression method with $\ell_1$ and $\ell_2$ regularization.  
    \item \textit{Remurs} \cite{song2017multilinear}: It is a multilinear regression model with $\ell_1$ norm and Tucker rank-based SNN regularization.
    \item \textit{Multi-way Multi-level Kernel Modeling (MMK)}  \cite{he2017multi}: It is a kernelized CP tensor factorization method to learn nonlinear features from tensors. Gaussain RBF kernel MMK is used with pre-computed kernel SVM.
\end{itemize}
SVM, Lasso, and ENet take vectorized fMRI data as input while Remurs and MMK directly take 3-D fMRI tensors as input. Lasso, ENet, Remurs, and Sturm can also be used for (embedded) feature selection. Therefore, we can add an SVM after each of them to obtain Lasso + SVM, ENet + SVM, Remurs + SVM and Sturm + SVM. The code for Sturm is built on the software library from \cite{lu2016libadmm,lu2018unified}. Remurs, Lasso, and ENet are implemented with the SLEP package \cite{liu2009slep}. MMK code is kindly provided by the first author of \cite{he2017multi}.

\begin{table*}[t!]
\centering
\begin{small}
\setlength{\tabcolsep}{4pt}
\begin{tabular}{m{2cm} m{1.9cm}m{1.9cm}m{1.9cm}m{1.9cm}m{1.1cm}m{1.1cm}m{1.1cm}}
\toprule
\multirow{2}{2cm}{ \textbf{Method}} & \multirow{2}{1.9cm}{\textbf{Resting 1}} & \multirow{2}{1.9cm}{\textbf{Resting 2}}  & \multirow{2}{1.9cm}{\textbf{Task 1}} & \multirow{2}{1.9cm}{\textbf{Task 2}} & \multicolumn{3}{c}{\textbf{Average}}\\\cline{6-8}
&&&&&\textbf{Resting} & \textbf{Task} & \textbf{All} \\
\midrule
SVM & 60.78 $\pm$ 0.09 & 63.97 $\pm$ 0.09 & \underline{87.38 $\pm$ 0.12} & 82.56 $\pm$ 0.17 & 62.38 & 84.97 & 73.67 \\
Lasso & 61.16 $\pm$ 0.08 & \underline{64.84 $\pm$ 0.11} & \underline{87.38 $\pm$ 0.12} & 85.22 $\pm$ 0.07 & \underline{63.00} & \underline{86.30} & \underline{74.65} \\
ENet & 61.21 $\pm$ 0.10 & 64.38 $\pm$ 0.10 & 81.19 $\pm$ 0.15 & 82.56 $\pm$ 0.17 & 62.80 & 81.87 & 72.34 \\
Remurs & 60.72 $\pm$ 0.08 & 62.13 $\pm$ 0.09 & 87.14 $\pm$ 0.13 & \underline{84.67 $\pm$ 0.15} & 61.43 & 85.90 & 73.67 \\
Sturm & 62.05 $\pm$ 0.11 & 63.47 $\pm$ 0.07 & \textbf{89.10 $\pm$ 0.09} & \textbf{86.89 $\pm$ 0.16} & 62.76 & \textbf{88.00} & \textbf{75.38} \\
\hline
Lasso + SVM & 63.37 $\pm$ 0.08 & 62.56 $\pm$ 0.09 & 74.05 $\pm$ 0.20 & 72.11 $\pm$ 0.16 & 62.97 & 73.08 & 68.02 \\
ENet + SVM & 64.20 $\pm$ 0.07 & 61.61 $\pm$ 0.08 & 76.43 $\pm$ 0.14 & 72.00 $\pm$ 0.14 & 62.91 & 74.21 & 68.56 \\
Remurs + SVM & \textbf{64.67 $\pm$ 0.10} & 60.23 $\pm$ 0.10 & 81.19 $\pm$ 0.12 & 83.56 $\pm$ 0.19 & 62.45 & 82.37 & 72.41 \\
Sturm+ SVM & \underline{64.66 $\pm$ 0.12} & \textbf{66.24 $\pm$ 0.06} & 78.10 $\pm$ 0.22 & 82.44 $\pm$ 0.16 & \textbf{65.45} & 80.27 & 72.86\\
\bottomrule
\end{tabular}
\end{small}
\caption{Classification accuracy (mean $\pm$ standard deviation in \%). 
Resting 1 and Resting 2 denote two disease diagnosis problems on ABIDE\textsubscript{NYU\&UM} and ADHD-200, respectively. Task 1 and Task 2 denote two neural decoding problems on OpenfMRI datasets for Balloon vs Mixed gamble and Simon vs Flanker, respectively. The best accuracy among all of the compared algorithms for each column is highlighted in \textbf{bold} and the second best is \underline{underlined}.}
\label{tab:classificationaccuracy}
\end{table*}

\begin{table*}[t!]
\centering
\begin{small}
\setlength{\tabcolsep}{4pt}
\begin{tabular}{m{2cm} m{1.9cm}m{1.9cm}m{1.9cm}m{1.9cm}m{1.1cm}m{1.1cm}m{1.1cm}}
\toprule
\multirow{2}{2cm}{ \textbf{Method}} & \multirow{2}{1.9cm}{\textbf{Resting 1}} & \multirow{2}{1.9cm}{\textbf{Resting 2}}  & \multirow{2}{1.9cm}{\textbf{Task 1}} & \multirow{2}{1.9cm}{\textbf{Task 2}} & \multicolumn{3}{c}{\textbf{Average}}\\\cline{6-8}
&&&&&\textbf{Resting} & \textbf{Task} & \textbf{All} \\
\midrule
Lasso & 0.52 $\pm$ 0.09 & 0.23 $\pm$ 0.32 & 0.74 $\pm$ 0.12 & 0.73 $\pm$ 0.01 & 0.38 & 0.73 & 0.55 \\
ENet & 0.60 $\pm$ 0.01 & 0.01 $\pm$ 0.01 & \textbf{0.96 $\pm$ 0.05} & \textbf{0.95 $\pm$ 0.03} & 0.31 & \textbf{0.96} & 0.63 \\
Remurs & 0.69 $\pm$ 0.03 & 0.73 $\pm$ 0.17 & 0.81 $\pm$ 0.08 & \underline{0.81 $\pm$ 0.07} & 0.71 & \underline{0.81} & \underline{0.76} \\
Sturm & \underline{0.86 $\pm$ 0.18} & \underline{0.86 $\pm$ 0.24} & 0.72 $\pm$ 0.24 & 0.60 $\pm$ 0.15 & \underline{0.86} & 0.66 & \underline{0.76} \\
\hline
Lasso + SVM & 0.57 $\pm$ 0.05 & 0.19 $\pm$ 0.40 & 0.77 $\pm$ 0.10 & 0.75 $\pm$ 0.06 & 0.38 & 0.76 & 0.57 \\
ENet + SVM & 0.58 $\pm$ 0.09 & 0.02 $\pm$ 0.01 & \textbf{0.96 $\pm$ 0.04} & \textbf{0.95 $\pm$ 0.04} & 0.30 & \textbf{0.96} & 0.63 \\
Remurs + SVM & 0.70 $\pm$ 0.13 & 0.74 $\pm$ 0.17 & 0.80 $\pm$ 0.04 & 0.79 $\pm$ 0.13 & 0.72 & 0.79 & 0.76 \\
Sturm + SVM & \textbf{0.87 $\pm$ 0.07} & \textbf{0.99 $\pm$ 0.01} & \underline{0.85 $\pm$ 0.14} & 0.56 $\pm$ 0.11 & \textbf{0.93} & 0.71 & \textbf{0.82}\\
\bottomrule
\end{tabular}
\end{small}
\caption{Sparsity (mean $\pm$ standard deviation) for respective results in Table \ref{tab:classificationaccuracy} with the \textbf{best} and \underline{second best} highlighted.}
\label{tab:sparsity}
\end{table*}

\subsection{Algorithm and evaluation settings}
\textbf{Model hyperparameter tuning.} Default settings are used for all existing algorithms. For Sturm, we follow the Remurs default setting \cite{song2017multilinear} to set $\rho$ to 1 and use the same set $\{10^{-3}, 5\times 10^{-3}, 10^{-2}, \dots, 5\times10^2, 10^3\}$ for  $\tau$ and $\gamma$, while scaling the first term in Eq. (\ref{eq.sturm.formulation}) by a factor $\alpha = \sqrt{(\max(I_1, I_2)\times I_3)}$ to better balance the scales of the loss function and regularization terms \cite{lu2016tensor,lu2018tensor}.

\textbf{Image resizing.} To improve computational efficiency and reduce the small sample size problem (and overfitting), the input 3-D tensors are further re-sized into three different sizes with a factor $\beta$, choosing from $\{0.3, 0.5, 0.7\}$.

\textbf{Feature selection.} In Lasso + SVM, ENet + SVM, Remurs + SVM, and Sturm + SVM, we rank the selected features by their associated absolute values of $\tens{W}$ in the descending order and feed the top $\eta \%$ of the features to SVM. We study five values of $\eta$: $\{1, 5, 10, 50, 100\}$.

\textbf{Evaluation metric and method.} The classification accuracy is our primary evaluation metric, and we also examine the sparsity of the obtained solution for all algorithms except SVM and MMK. For a particular binary classification problem, we perform ten-fold cross validation and report the mean and standard deviation of the classification accuracy and sparsity over ten runs. For each of the ten (test) folds, we perform an inner nine-fold cross validation using the remaining nine folds to determine $\tau$ and $\gamma$ (jointly for ENet, Remurs, and Sturm on a $13 \times 13$ grid), $\beta$, and $\eta$ above that give the highest classification accuracy, with the corresponding sparsity recorded. The sparsity is calculated as the ratio of the number of zeros in the output coefficient tensor $\tW$ to its size $I_1\times I_2 \times I_3$. In general, higher sparsity implies better interpretability \cite{hastie2015statistical}.

\subsection{Results and discussion}

Table \ref{tab:classificationaccuracy} reports the classification accuracy for all algorithms except MMK. This is because all MMK results are below 60\% on all four problems, possibly due to the default settings of the CP rank and SVM kernel. (We have tried a few alternative settings without improvement, though further tuning may still lead to better results.) Table \ref{tab:sparsity} presents the respective sparsity values except SVM, which uses all features so the sparsity is zero. In both tables, the best results are highlighted in \textbf{bold}, with the second best ones \underline{underlined}. 

\textbf{Performance on Resting 1 \& 2.} For these two resting-state problems, Sturm + SVM has the highest accuracy of 65.45\%, and Lasso is the second best with 2.45\% lower accuracy. In terms of sparsity, Sturm + SVM and Sturm are the top two. Specifically, on Resting 1, Remurs + SVM and Sturm + SVM are the top two algorithms with almost identical accuracy of 64.67\% and 64.66\%, respectively. Moreover, Sturm + SVM also has the highest sparsity of 0.87 and Sturm has the second-highest sparsity of 0.86. For Resting 2, Sturm + SVM has outperformed all other algorithms on both accuracy (66.24\%) and sparsity (0.99). 

\textbf{Performance on Task 1 \& 2.}
For these two task-based problems, Sturm has outperformed all other algorithms in accuracy, with 89.10\% on Task 1 and 86.89\% on Task 2. Lasso is again the second best in accuracy. ENet and ENet + SVM has the best sparsity of 0.96 while their accuracy values are only 81.87\% and 74.21\%, respectively. Sturm + SVM has significant drop in accuracy compared with Sturm alone, and Lasso + SVM, ENet + SVM and Remurs + SVM all have lower accuracy compared to without SVM.

\textbf{Summary.} There are four key observations on the whole:
\begin{itemize}
    \item Sturm has the best overall accuracy of 75.38\%. Sturm has outperformed Remurs in accuracy for all four classification problems. The only difference between Sturm and Remurs is replacing SNN with TNN. Therefore, this superiority indicates that tubal rank-based TNN is superior to Tucker rank-based SNN in the supervised, regression setting.
    \item The reported sparsity corresponds to the best solution determined via nine-fold cross validation. Lasso and Lasso + SVM have the lowest, i.e., poorest, sparsity. ENet and ENet + SVM also have much lower sparsity than Remurs/Sturm and their + SVM versions, more than 0.10 (10\%) lower. On the other hand, ENet and ENet + SVM have the highest sparsity on task-based fMRI while getting a solution closely to zero sparsity on Resting 2, showing high variation.
    \item Performing SVM after the four regression methods can improve the classification accuracy (though not always) on resting-state fMRI, while it has degraded their classification performance on task-based fMRI in all cases. In particularly, Lasso + SVM and Sturm + SVM on Task 1, and Lasso + SVM and ENet + SVM on Task 2 have dropped more than 10\% in accuracy.
    \item Disease diagnosis on resting-state fMRI is significantly more challenging than neural decoding on task-based fMRI. Although it should be aware from Table \ref{tab:dataset} that the number of samples is different for the resting-state and task-based fMRI, engaged brain activities are generally easier to classify than those not engaged and the difference is consistent with results reported in the literature.
\end{itemize} 

\subsection{Analysis}
\begin{figure}[t]
    \centering
     \begin{subfigure}[b]{0.235\textwidth}
         \includegraphics[width=\linewidth]{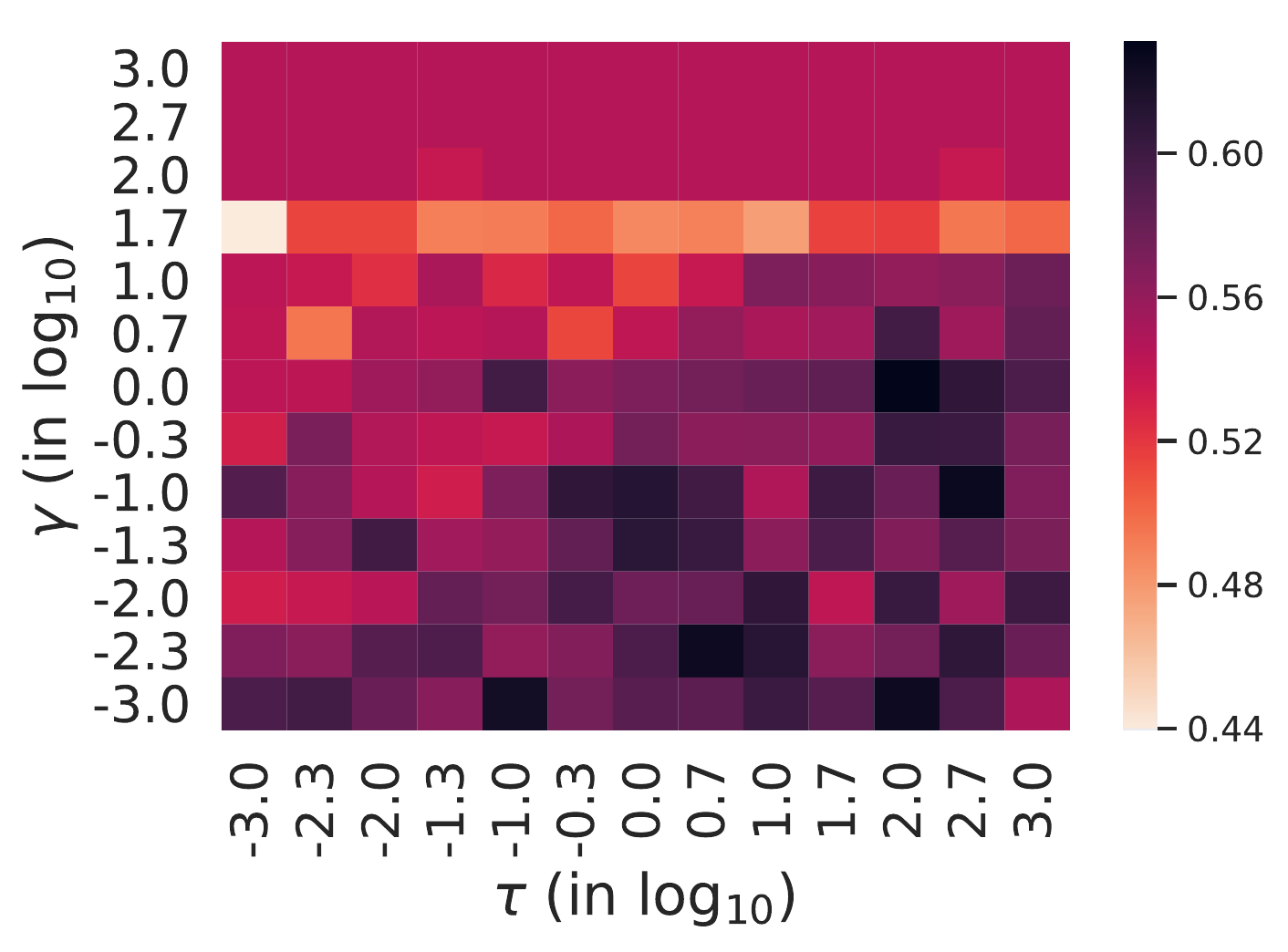}
         \caption{{Resting 2}}
         \label{fig:ADHDSensitivity}
    \end{subfigure}
    \begin{subfigure}[b]{0.235\textwidth}
    \includegraphics[width=\linewidth]{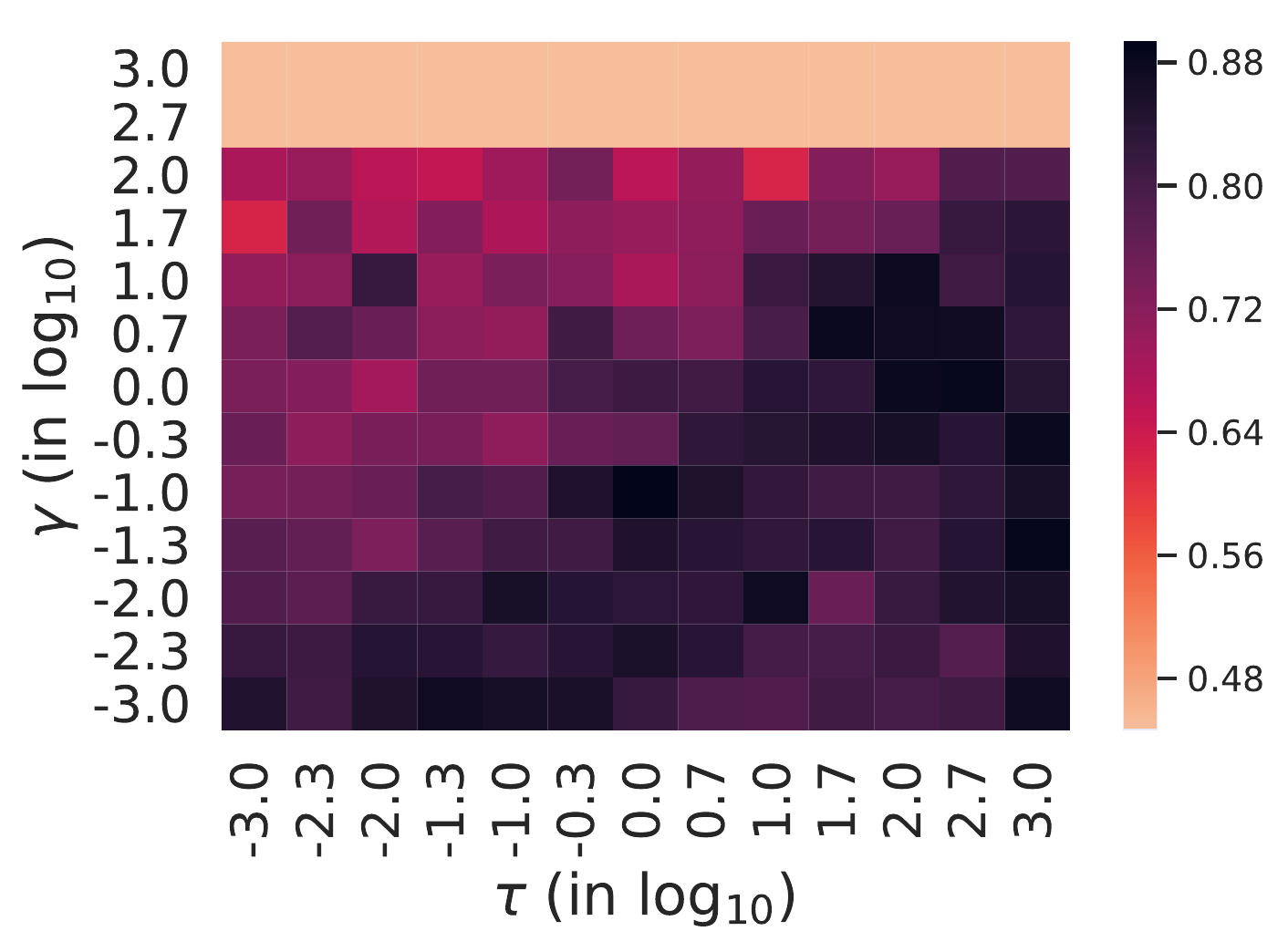}
    \caption{{Task 2}}
      \label{fig:OpenfMRISensitivity}
    \end{subfigure}
    \caption{Sensitivity study on hyperparameters $\tau$ and $\gamma$. Darker color indicates better classification accuracy (best viewed on screen).}
    \label{fig:hyperparametersensitivity}
\end{figure}

\textbf{Hyperparameter sensitivity.} Figure~\ref{fig:hyperparametersensitivity} illustrates the classification performance sensitivity of Sturm on its two hyperparameters $\tau$ and $\gamma$ for two problems: Resting 2 and Task 2. In general, the lower right, i.e., large $\tau$ and  small $\gamma$ values, shows higher accuracy. This implies that the tubal rank regularization helps more in improving the accuracy than sparsity regularization. However, Fig. \ref{fig:ADHDSensitivity} shows poorer smoothness than Fig. \ref{fig:OpenfMRISensitivity}, another indication of resting-state fMRI being more challenging than task-based fMRI. This makes hyperparameter tuning more difficult for resting-state fMRI, (partly) causing the poorer classification performance than task-based fMRI.

\begin{figure}[t]
    \centering
     \begin{subfigure}[b]{0.235\textwidth}
         \includegraphics[width=\linewidth]{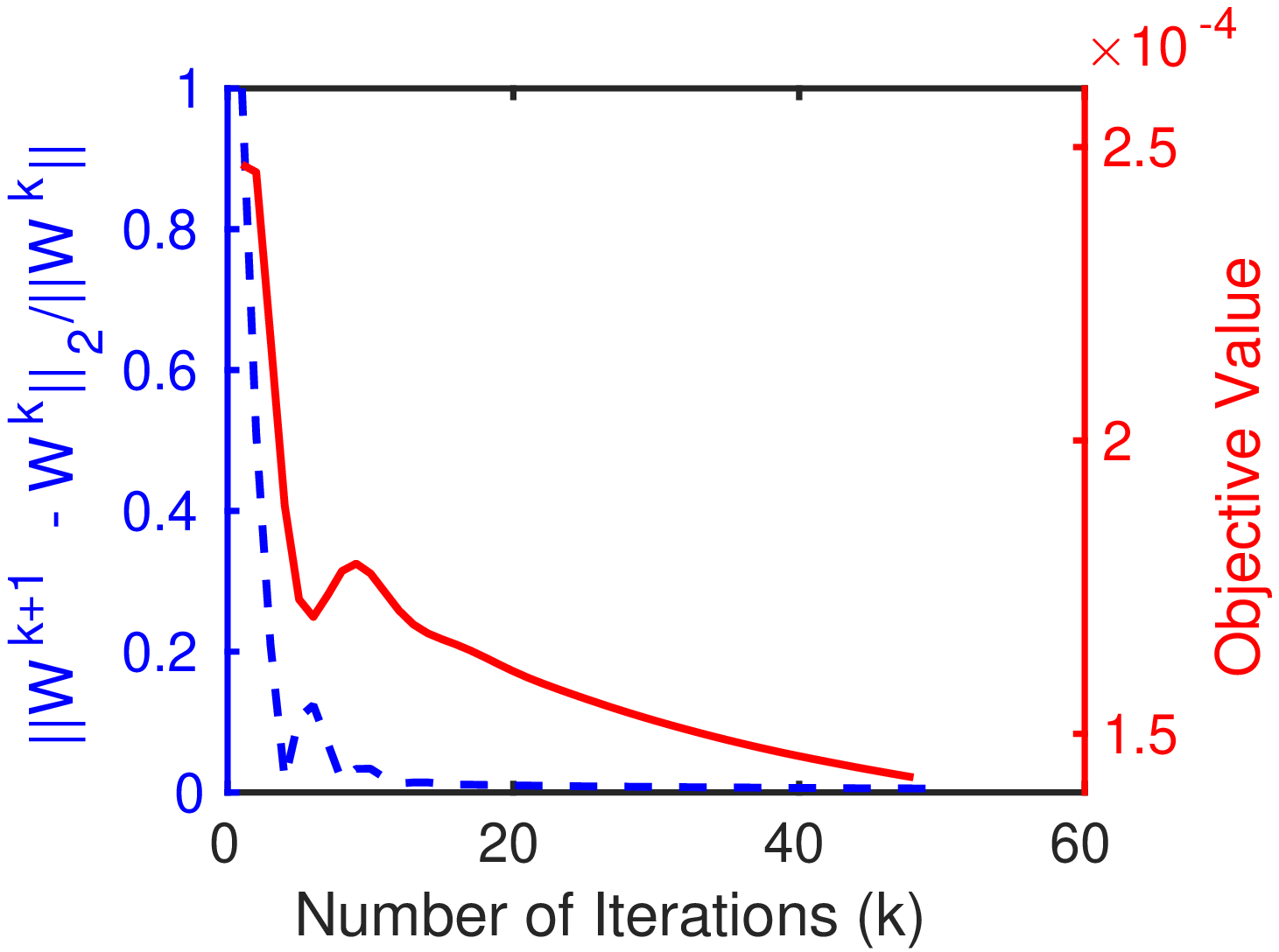}
         \caption{{Resting 2}}
         \label{fig:ADHDConvergence}
    \end{subfigure}
    \begin{subfigure}[b]{0.235\textwidth}
    \includegraphics[width=\linewidth]{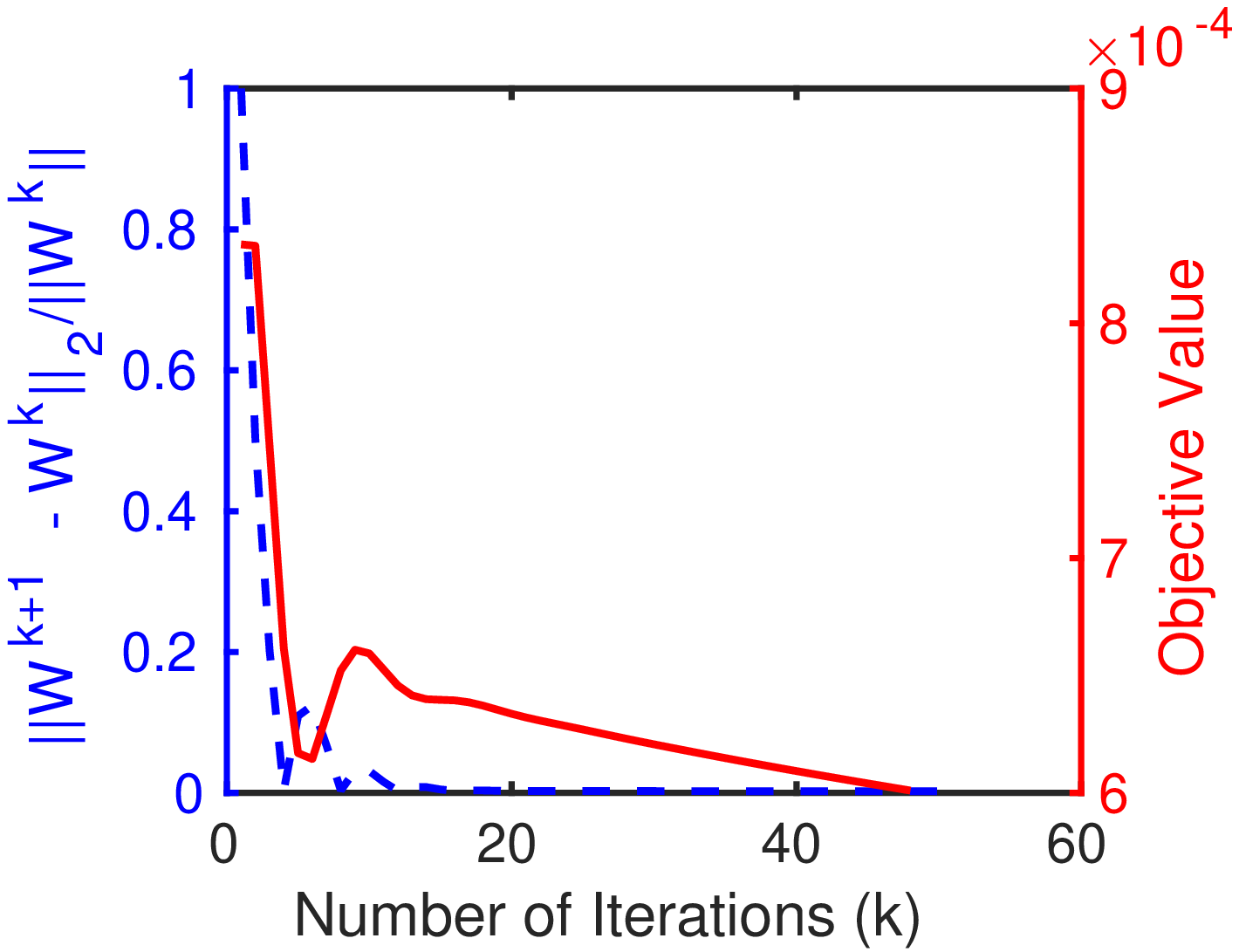}
    \caption{{Task 2}}
      \label{fig:OpenfMRIT21T22Convergence}
    \end{subfigure}
    \caption{Convergence analysis (best viewed on screen).}
    \label{fig:convergenceAnalysis}
\end{figure}

\textbf{Convergence analysis.} Figure \ref{fig:convergenceAnalysis} shows the convergence of $\tens{W}$ and the Sturm objective function value in (\ref{eq.sturm.formulation}) on Resting 2 and Task 2. It can be seen that $\tens{W}$ has a fast convergence speed in both cases, though the objective function converges at a slower rate. 

\section{Conclusion}
In this paper, we proposed a sparse tubal rank regularized multilinear regression model (\textit{Sturm}). It performs regression with penalties on the tubal tensor nuclear norm, a convex relaxation of the tubal rank, and the standard $\ell_1$ norm. To the best of our knowledge, this is the first \textit{supervised} learning method using TNN. The Sturm algorithm is derived via a standard ADMM framework.

We evaluated Sturm (and Sturm + SVM) in terms of classification accuracy and sparsity against eight other methods on four binary classification problems from three public fMRI repositories. The datasets include both resting-state and task-based fMRI, unlike most existing works focusing on only one of them. The results showed the superior overall performance of Sturm (and Sturm + SVM in some cases) over other methods and confirmed the benefits of TNN and tubal rank regularization in a supervised setting. 

{\small
\bibliographystyle{icml2018}
\bibliography{egbib}
}

\end{document}

%% file: notations.tex
\newcolumntype{C}[1]{>{\centering\let\newline\\\arraybackslash\hspace{0pt}}m{#1}}

\theoremstyle{definition}

\makeatletter
\newsavebox\myboxA
\newsavebox\myboxB
\newlength\mylenA

\newcommand*\xbar[2][0.75]{%
    \sbox{\myboxA}{$\m@th#2$}%
    \setbox\myboxB\null
    \ht\myboxB=\ht\myboxA%
    \dp\myboxB=\dp\myboxA%
    \wd\myboxB=#1\wd\myboxA
    \sbox\myboxB{$\m@th\overline{\copy\myboxB}$}
    \setlength\mylenA{\the\wd\myboxA}
    \addtolength\mylenA{-\the\wd\myboxB}%
    \ifdim\wd\myboxB<\wd\myboxA%
       \rlap{\hskip 0.5\mylenA\usebox\myboxB}{\usebox\myboxA}%
    \else
        \hskip -0.5\mylenA\rlap{\usebox\myboxA}{\hskip 0.5\mylenA\usebox\myboxB}%
    \fi}
\makeatother

\newcommand{\tens}[1]{\mathcal{#1}}
\newcommand{\vect}[1]{\ensuremath{\mathbf{#1}}}
\newcommand{\mat}[1]{\ensuremath{\mathbf{#1}}}

\newcommand{\R}{\mathbb{R}}

\newcommand{\tA}{\tens{A}}
\newcommand{\tT}{\tens{T}}
\newcommand{\tB}{\tens{B}}
\newcommand{\tX}{\tens{X}}

\newcommand{\tU}{\tens{U}}
\newcommand{\tV}{\tens{V}}
\newcommand{\tS}{\tens{S}}

\newcommand{\tW}{\tens{W}}
\newcommand{\tP}{\tens{P}}
\newcommand{\tQ}{\tens{Q}}

\newcommand{\ftA}{\tA_\mathscr{F}}
\newcommand{\ftT}{\tT_\mathscr{F}}

\newcommand{\fmA}{\A_\mathscr{F}}
\newcommand{\fmT}{\mat{T}_\mathscr{F}}
\newcommand{\ftZ}{\tens{Z}_\mathscr{F}}
\newcommand{\fmZ}{\mat{Z}_\mathscr{F}}

\newcommand{\la}{\langle}
\newcommand{\ra}{\rangle}

\newcommand{\A}{\mat{A}}

\newcommand{\I}{\mat{I}}

\newcommand{\U}{\mat{U}}
\newcommand{\V}{\mat{V}}

\newcommand{\X}{\mat{X}}

\newcommand{\mA}{\mat{A}}

\newcommand{\w}{\vect{w}}

\newcommand{\y}{\vect{y}}